%% file: main.tex
\documentclass[12pt, anon]{l4dc2024}

\title[Parametric FaSTrack]{Parameterized Fast and Safe Tracking (FaSTrack) using DeepReach}
\usepackage{times}
\input{l4dc_2024/notation}

\input{l4dc_2024/packages}

\author{%
 \Name{{Hyun Joe} Jeong} \Email{hjjeong@ucsd.edu}\\
 \addr Department of Mechanical and Aerospace Engineering, University of California San Diego
 \AND
 \Name{Zheng Gong} \Email{zhgong@ucsd.edu}\\
 \addr Department of Mechanical and Aerospace Engineering, University of California San Diego
 \AND
 \Name{Somil Bansal} \Email{somilban@usc.edu}\\
 \addr Department of Electrical and Computer Engineering, University of Southern California
 \AND
 \Name{Sylvia Herbert} \Email{sherbert@ucsd.edu}\\
 \addr Department of Mechanical and Aerospace Engineering, University of California San Diego%
}

\begin{document}

\maketitle

\begin{abstract}%
 Fast and Safe Tracking (FaSTrack, \cite{Herbert2017}) is a modular framework that provides safety guarantees while planning and executing trajectories in real time via value functions of Hamilton-Jacobi (HJ) reachability. These value functions are computed through dynamic programming, which is notorious for being computationally inefficient. Moreover, the resulting trajectory does not adapt online to the environment, such as sudden disturbances or obstacles.
 DeepReach (\cite{bansal2020deepreach}) is a scalable deep learning method to HJ reachability that allows parameterization of states, which opens up possibilities for online adaptation to various controls and disturbances. In this paper, we propose \textit{Parametric FaSTrack}, which uses DeepReach to approximate a value function that parameterizes the control bounds of the planning model. The new framework can smoothly trade off between the navigation speed and the tracking error (therefore maneuverability) while guaranteeing obstacle avoidance in a priori unknown environments. We demonstrate our method through two examples and a benchmark comparison with existing methods, showing the safety, efficiency, and faster solution times of the framework.
\end{abstract}




\section{Introduction}
\input{sections/introduction}
\section{Background}\label{sec:background}
\input{sections/background}
\section{Offline Parameterized Value Function Training}\label{sec:formulation}
\input{sections/formulation}

\section{Online Adaptive Safe Planning and Control}\label{sec:planning}
\input{sections/planning}

\section{Results}\label{sec:results}
\input{sections/results}

\section{Conclusions}\label{sec:conclusions}
\input{sections/conclusions}


\bibliography{l4dc_2024/references}

\end{document}

%% file: l4dc_2024/notation.tex
\newcommand{\shnote}[1]%
    {\textcolor{magenta}{ #1}}

\newcommand{\pset}{\mathcal{P}} 
\newcommand{\tset}{\mathcal{S}} 
\newcommand{\tvar}{t}
\newcommand{\thor}{t_f} 
\newcommand{\tstate}{s} 
\newcommand{\pstate}{p} 

\newcommand{\rstate}{r} 

\newcommand{\ttraj}{\xi_{\tdyn}} 
\newcommand{\rtraj}{\xi}

\newcommand{\tctrl}{u_s} 
\newcommand{\dstb}{d} 
\newcommand{\pctrl}{u_p} 
\newcommand{\tdyn}{f} 
\newcommand{\pdyn}{h} 
\newcommand{\rdyn}{g} 
\newcommand{\dset}{\mathcal{D}}
\newcommand{\tcset}{\mathcal{U}_s} 
\newcommand{\tcfset}{\mathbb{U}_s} 
\newcommand{\pcset}{\mathcal{W}} 
\newcommand{\dfset}{\mathbb{D}}
\newcommand{\ptmat}{Q} 

\newcommand{\valfunc}{V} 
\newcommand{\minval}{\underline{V}} 
\newcommand{\TEB}{\mathcal B} 

\newcommand{\rtrans}{\Phi}

\newcommand{\dTEB}{\TEB_{e,d}^\infty(\beta)}
\newcommand{\dTEBmax}{\mathcal{K}}
\newcommand{\dist}{\mathcal{D}}



%% file: l4dc_2024/packages.tex
\usepackage[utf8]{inputenc} 
\usepackage[T1]{fontenc}    
\usepackage{hyperref}       
\usepackage{url}            
\usepackage{booktabs}       
\usepackage{amsfonts}       
\usepackage{nicefrac}       
\usepackage{microtype}      
\usepackage{xcolor}         
\usepackage{wrapfig}
\usepackage{graphicx} 
\usepackage{graphicx} 
\usepackage{algorithm}
\usepackage{algpseudocode}
\usepackage{amsmath,amssymb}

\usepackage{bbm}
\usepackage{soul}
\usepackage{microtype,wrapfig}
\usepackage{multicol}

\usepackage[export]{adjustbox}
\usepackage{makecell}
\usepackage{caption}
\usepackage{wrapfig}

%% file: sections/introduction.tex
In order for autonomous dynamical systems to be viable in real world applications, they often require rigorous safety guarantees and efficient computation. Further, many environments will be \textit{a priori} unknown, adding to the threshhold of feasibility. 
Traditionally, real time navigation for autonomous systems starts with creating waypoints using a geometric or kinodynamic planner (\cite{Schulman2013}; \cite{Kobilarov2012}). Tracking controllers, such as model predictive control (MPC) (\cite{Qin2003}; \cite{Alexis2016}), are then used to track these waypoints. However, the geometric planners do not account for system dynamics, making them dynamically infeasible. This can have devastating effects when the system is near obstacles. 
MPC based methods (\cite{7487277,8558663}) are proposed which sample various trajectories for efficient trajectory planning. However, no safety guarantees are involved, which can lead to safety violations.
To generate safety guarantees for nonlinear systems with control bounds, control barrier functions (\cite{choi2021robust,manjunath2021safe,tonkens2022refining, ahmad2022adaptive}) and Hamilton Jacobi (HJ) reachability analysis (\cite{fisac2015reach,bansal2017hamilton,9304186,9855830}) are popular methods for generating safe sets and controllers that minimally modify a reference controller. Although safe, online update of these methods is difficult in unknown environments.


Fast and Safe Tracking (\cite{Herbert2017}) is an modular framework built on HJ reachability analysis for guaranteed safe tracking of a general (possibly geometric) planning algorithm through an unknown environment. 
The safety guarantees stem from precomputing a tracking error bound (TEB) and controller that is \textit{robust to any planning algorithm}, even one that may act adversarially. Online, obstacles are augmented by this error bound, and the planning algorithm must seek a path using these augmented obstacles. Because safety is guaranteed with no assumptions on the specific planning algorithm or its behavior, this approach often results in very conservative trajectories. 
Planner aware FaSTrack (\cite{sahraeekhanghah2021pa}) and
Meta-FaSTrack (\cite{fridovich2018planning}) are proposed to overcome this limitation under additional assumptions on the planning algorithm. However, these methods are still fundamentally limited by the computational complexity of HJ reachability, which scales exponentially in the number of state dimension. Moreover, adapting the tracking error bound with changes in planner control or system uncertainties remains challenging.

\begin{figure}[h]
    \centering
    \includegraphics[width=.85\textwidth]{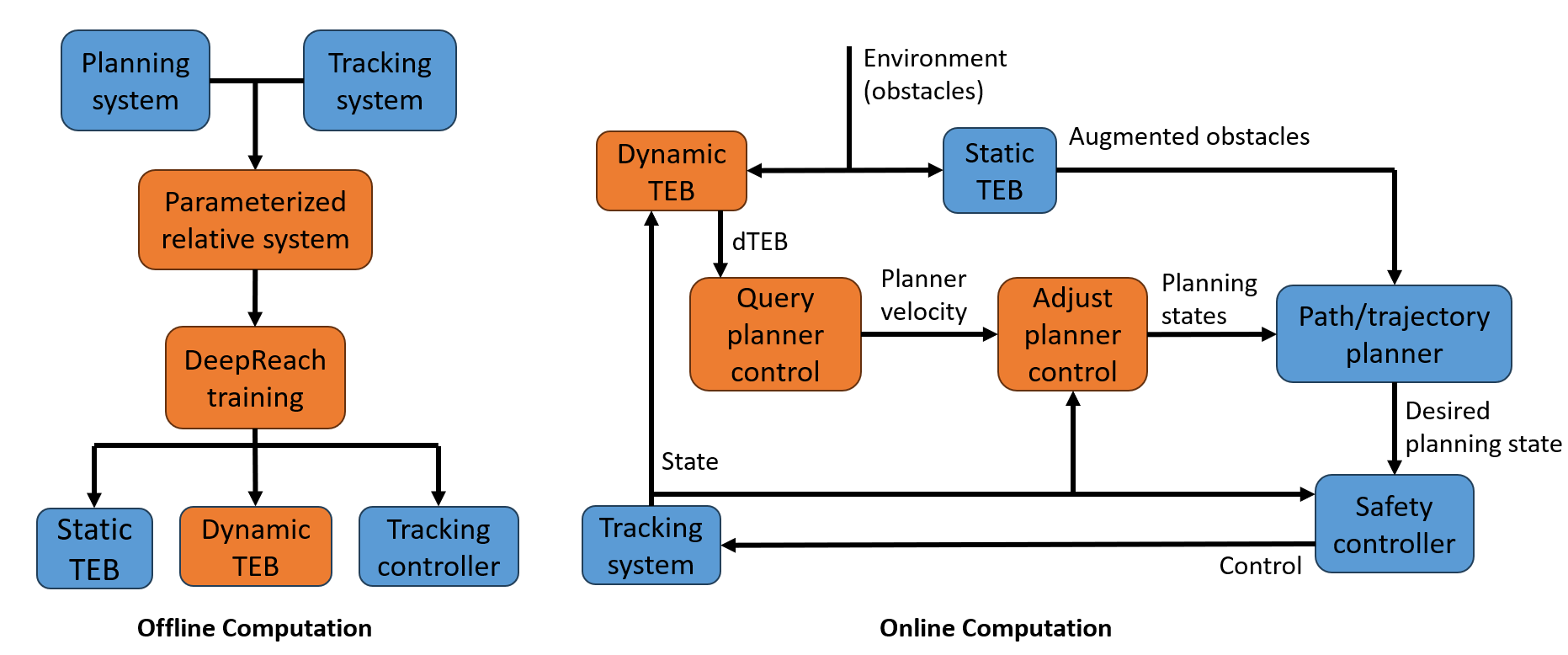}
    \caption{Left figure denotes the offline framework (performed once), and the right figure denotes the online framework (performed every iteration). Components of FaSTrack are shown in blue, while components of PF are shown in orange.\vspace{-1.5em}}
    \label{fig:1}
\end{figure}

DeepReach (\cite{bansal2020deepreach}) is a neural PDE solver for high-dimensional HJ reachability analysis. This approach provides safe sets and controllers order of magnitudes faster than standard methods, while providing probabilistic safety assurances. Many approaches catalyzed by DeepReach has been proposed, such as parameterizing the value function for online adaptation (\cite{10160554}), and parameterizing input bounds for real time adaptation to environmental and system uncertainties (\cite{10160828}). We combine DeepReach with FaSTrack to propose the Parametric FaSTrack (PF) framework (Fig.~\ref{fig:1}). Our key contributions are as follows:
\begin{enumerate}
    \item We use DeepReach to approximate the HJ value function (and the corresponding TEB and controller), which improves scalability to high dimensional systems. \vspace{-2mm}
    \item The TEB and controller are \textit{parameterized by the speed of the planning algorithm, allowing the planner to automatically trade off between safety and effiency}. For instance, In open environments, augmenting obstacles by a large TEB associated with fast planning is acceptable. In cluttered environments, slower planning and tighter TEB is required for safety. \vspace{-2mm}
    \item We provide algorithms that smoothly switches between different TEB to trade off between safety and efficiency. The overall navigation speed is increased by 40\%, as shown in the examples compared to state-of-the-art online planning methods. \vspace{-2mm}
\end{enumerate}

%% file: sections/background.tex
In this section we introduce the FaSTrack algorithm for safe navigation and DeepReach for high-dimensional reachability analysis.  

\subsection{Fast and Safe Tracking (FaSTrack) Framework}



\noindent \textbf{Tracker Model. }In the FaSTrack framework, the \textit{tracker model} represents the true robot and the \textit{planner model} represents the real-time path planner. 
The \textit{tracker model} is given by the following ordinary differential equation (ODE)
\begin{align} \label{eq:tdyn}
    \frac{d \tstate}{d\tvar} =\dot \tstate  = \tdyn ( \tstate (\tvar),\tctrl (\tvar),\dstb (\tvar)  ), \quad \tvar \in [0,\thor], \quad \tstate \in \tset \subset \mathbb R^{n_s},\tctrl \in \tcset \subset \mathbb R^{n_u},\dstb \in \dset \subset \mathbb R^{n_d},
\end{align}
where $\tvar$ is the time and $\tstate$, $\tctrl$ and $\dstb$ are the tracker state, control and disturbance at a given time respectively, with $\tcset$ and $\dset$ to be compact sets. $\tctrl (\cdot) \in \tcfset$ and $\dstb (\cdot) \in \dfset$ are the control and disturbance signal, assumed to be measurable: $\tctrl (\cdot): [0,\thor]\mapsto \tcset$ and $\dstb (\cdot): [0,\thor]\mapsto \dset$.
We further assume the dynamics $\tdyn : \tset \times \tcset \times \dset \mapsto \tset$ is Lipschitz continuous in $\tstate$ for fixed $\tctrl (\cdot)$ and $\dstb (\cdot)$. A unique solution of \eqref{eq:tdyn} can be solved as $\ttraj (t; 0, \tstate, \tctrl (\cdot),\dstb(\cdot))$, or in short $\ttraj (t)$. 

\noindent \textbf{Planner Model.} The \textit{planner model} is a virtual model designed for path planning, given ODE 
\begin{align*}
    \frac{d \pstate}{ dt} = \dot \pstate = \pdyn(\pstate,\pctrl), \quad \tvar \in [0,\thor], \quad \pstate \in \pset \subset \mathbb R^{n_p}, \pctrl \in \pcset,
\end{align*}
where $\pstate$ is the planner state and $\pctrl$ is the planner control. We assume the planner state is a subset of the tracker state, and the planner space is a subspace of the tracker space.

Since path planning is done using the planner model, where goal and constraints sets in the planner space are denoted as $\mathcal G_p$ and $\mathcal C_p$, respectively. Also, since we assume unknown environment and limited sense range, we denote the sensed environment as $\mathcal C_{p,\text{sense}}(t)$, and the augmented obstacle as $\mathcal C_{p,\text{aug}}(t)$. Note both $\mathcal C_{p,\text{sense}}(t)$ and $\mathcal C_{p,\text{aug}}(t)$ are time-varying sets that update at each iteration.  

\noindent \textbf{Relative Dynamics.} The FaSTrack framework provides guarantees by precomputing the maximum error (i.e. distance) possible between the tracker model and planner model. In order to compute this, the \textit{relative dynamics} between the two systems must be determined.
Since the planner space is a subspace of the tracker space, we can always find a matrix $\ptmat \in \mathbb R^{n_s \times n_p}$, and assume we can also find a linear tranformation $\rtrans(\tstate,\pstate)$ s.t. $\rstate = \rtrans(\tstate,\pstate) (\tstate -\ptmat \pstate)$ and 
\begin{align*}
    \dot \rstate = \rdyn (\rstate, \tctrl,\pctrl, \dstb), 
\end{align*}
with solution $\rtraj (t;0,\rstate, \tctrl, \pctrl ,\dstb)$. We denote the common states in tracker and planner as $e$ and the rest as $\eta$, i.e. $\rstate = [e,\eta]$. 

\vspace{2mm}
\noindent \textit{Running example: }consider the tracker model as a Dubin's car with controls $\omega$ and $\alpha$:  
\begin{align*}
    \dot  \tstate_1 = \tstate_4 \sin (\tstate_3), \hspace{2mm}
    \dot  \tstate_2 = \tstate_4 \cos (\tstate_3), \hspace{2mm}
    \dot  \tstate_3 = \omega, \hspace{2mm}
    \dot  \tstate_4 = \alpha.
\end{align*}
Consider the planning model: 
$\dot  \pstate_1 = u_{px}, \hspace{2mm}
    \dot  \pstate_2 = u_{py},$
where $u_{px}$ and $u_{py}$ are planning controls. Given linear transformation $\rtrans = I_4$ and matrix $\ptmat = \begin{bmatrix} 1&0&0&0 \\ 0&1&0&0\end{bmatrix}^T$, the relative model is:
\begin{align} \label{eq:ex_dubins}
    \dot \rstate_1 = \rstate_4 \sin (\rstate_3) - u_{px}, \hspace{2mm}\dot \rstate_2 = \rstate_4 \cos (\rstate_3 ) - u_{py}, \hspace{2mm}\dot \rstate_3 = \omega, \hspace{2mm}\dot \rstate_4 = \alpha.
\end{align}


\noindent \textbf{Cost Function.} With the relative dynamics in hand, the maximum possible relative distance (i.e. TEB) between the two models can be computed.  The error function of this game is denoted as $\ell (\rstate)$, and is chosen to be the Euclidean norm in the relative state space (here, low error corresponds to accurate tracking). We assume that the tracker model seeks to minimize this tracking error by \textit{pursuing} the planner model. 
 Because the environment and behavior of the planner model is unknown \textit{a priori}, to provide safety guarantees we must assume worst-case actions of the planner model.  This amounts to assuming the planner model seeks to maximize the tracking error and \textit{evade} the tracker model. We have therefore constructed a pursuit-evasion game between the tracker and planner. 
 
 \noindent \textbf{Offline Computation of the Tracking Error Bound and Controller.} We define the strategies of the planner and disturbance as mappings: $\gamma _p : \tcset \mapsto \pcset$, $\gamma _d : \tcset \mapsto \dset$ and assume they are restricted to be the non-anticipative strategies $\gamma _p \in \Gamma_p$, $\gamma _d \in \Gamma_d$. The value of the game can be defined as: 
\begin{align} \label{eqn:value function}
    \valfunc (\rstate, \thor) = \sup _{\gamma _p \in \Gamma_p, \gamma _d \in \Gamma_d} \inf _{u_s\in \tcfset} \{ \max_{t\in [0 , \thor]} \ell(\rtraj(t; 0, \rstate, \tctrl (\cdot), \gamma _p(\cdot),\gamma _d(\cdot))) \}.
\end{align}
It has been shown that \eqref{eqn:value function} can be computed by solving the following HJI-VI recursively: 
\begin{align}\label{eqn:HJIVI}
    0 = \max \{ \ell(\rstate) - \valfunc(\rstate, t), \quad \frac{\partial \valfunc}{ \partial t} + \min_{\tctrl\in \tcset} \max_{\pctrl \in \pcset,\dstb \in \dset}  \nabla V \cdot \rdyn(\rstate,\tctrl,\pctrl,\dstb) \}, \quad \valfunc(\rstate, 0) = \ell(\rstate),
\end{align}
where $H(t,r,\frac{\partial V}{\partial r}) = \min_{\tctrl\in \tcset} \max_{\pctrl \in \pcset,\dstb \in \dset}  \nabla V \cdot \rdyn(\rstate,\tctrl,\pctrl,\dstb)$ is the Hamiltonian. The value of at a particular $r,t_f$ corresponds to the largest tracking error, that can occur over the time horizon. The smallest non-empty level set of this function therefore provides the smallest TEB and the set of relative states for which this bound can be achieved over the time horizon.

In the case where the limit function $\valfunc ^\infty (\rstate) = \lim_{\thor \rightarrow \infty} \valfunc (\rstate, \thor)$ exists, every non-empty level set is a robust control invariant set.  The set at level $\minval^\infty = \min_{\rstate} \valfunc ^\infty (\rstate)$, is called the infinite-time TEB: the relative system will stay inside this set for all $t\geq 0$. Throughout the paper, we assume this limit function exist, though extensions to time-varying error bounds have been successful. The infinite-time TEB in the relative space and planner space are denoted by
\begin{align*}
    \TEB^\infty := \{ \rstate : \valfunc^\infty(\rstate) \leq \minval ^\infty \}, \quad \TEB^\infty _e := \{e: \exists \eta \text{ s.t. }\valfunc^\infty(e,\eta) \leq \minval ^\infty \}.
\end{align*}


\noindent \textbf{Online Planning and Tracking.} Online, the sensed obstacles are augmented by this error bound projected into the planner state space.  The planning algorithm employs the planning model to determine the next planner state.  The autonomous system (represented by the tracker model) can efficiently compute the optimal tracking control via the relative state between itself and the planner: 

\begin{equation}
    \tctrl^* = \text{argmin}_{\tctrl \in \tcset}\max_{\pctrl \in \pcset,\dstb \in \dset} \nabla V \cdot \rdyn(\rstate,\tctrl,\pctrl,\dstb). \label{eq:tracking_control}
\end{equation}
This process repeats until the system has reached the goal.



\subsection{DeepReach and Parameter-Conditioned Reachability}

While traditional reachability frameworks solves the HJI-VI over a grid, DeepReach approximates the value function by having a sinusoidal deep neural network (DNN) learn a parameterized estimation of the value function. The key benefit to this approach is that memory and complexity requirements for training are tied to the complexity of the value function rather than the grid resolution. DeepReach trains the DNN using self-supervision on the HJB-VI itself. By inputting state, time, and parameters, it outputs a learned value function.

Parameter-conditioned reachability (PCR) further parameterizes the reachable sets via uncertain environment parameters $\beta$. These parameters are then treated as a virtual state in the system model. Though this virtual state has no dynamics, it allows the DNN to learn a family of value functions simultaneously, producing a parameterized value function $V(x,\beta)$. PCR effectively diminishes separate training needs for value functions corresponding to the changed system, and solves them in the same fashion as DeepReach, which also allows for a swift computation time of high dimensional systems. 
A system can leverage the effectiveness of PCR online by taking into account any sudden changes in the parameters $\beta$ by just doing a simple query of the value function corresponding to the changed parameter value. Its online adaptability improves a system's robustness by maintaining its safety. The optimal control can also be determined from the gradients of the learned value function.

%% file: sections/formulation.tex
We parameterize the offline computation in FaSTrack with the planner control bound, and use DeepReach to approximate the value function that corresponds to the new relative dynamics. This portion outputs the static TEB (sTEB), dynamic TEB (dTEB), and the tracking controller, as shown in Fig.~\ref{fig:1} (left). The sTEB is the TEB given minimal planner control bound, while the dTEB is determined by the planner control bound used. 


\noindent \textbf{Parameterized Relative System.}
We start by parameterizing the planner control bound $\pcset (\beta)$: 
\begin{align*}
    \dot {\bar \rstate} =
    \begin{bmatrix}
        \dot \rstate \\ \dot \beta
    \end{bmatrix} = 
    \begin{bmatrix}
        \rdyn (\rstate, \tctrl, \pctrl, \dstb) \\ 0
    \end{bmatrix},
\end{align*}
where $\bar \rstate = [\rstate,\beta]$ is the augmented relative states, and $\beta $ is the virtual state with zero dynamics. We assume $\pctrl \in \pcset (\beta)$, and $\beta \in [\beta_l ,\beta_u]$. The Hamiltonian for the augmented relative dynamics is $H(\rstate,\nabla V_\theta,\tvar;\beta) = \min_{\tctrl\in \tcset} \max_{\pctrl \in \pcset(\beta),\dstb \in \dset} \nabla V_\theta \cdot \rdyn(\rstate,\tctrl,\pctrl,\dstb)$. Note that $V(\bar \rstate,\tvar) = V(r,\tvar;\beta)$.



For the running example, the first four states of the relative system are identical to \eqref{eq:ex_dubins}. The control bounds are given as $r_3 \in [-5, 5]$ and $r_4 \in [-1, 1]$. We add $\dot r_5 = \dot \beta _1 = 0$ and $\dot r_6 = \dot \beta _2 = 0$ as the virtual parameter states with no dynamics, with $\beta_1,\beta_2 \in [0.5 , 1.25]$:
\begin{align} \label{eqn: para Dubins}
    \dot r_1 = r_4 \sin (r_3) - u_{px}, \hspace{2mm}\dot r_2 = r_4 \cos (r_3) - u_{py}, \hspace{2mm}\dot r_3 = \omega, \hspace{2mm}\dot r_4 = \alpha, \hspace{2mm}\dot \beta_1 = 0, \hspace{2mm}\dot \beta_2 = 0.
\end{align}

\noindent \textbf{DeepReach Training.} The loss function $h$ used to train the DNN is given by 
\begin{align}\label{eq:loss_function} h_1(r_i,t_i,\beta_i;\theta) &= \left|\left|V_\theta(r_i,t_i;\beta_i) - l(r_i)\right|\right|_{(t_i = T)}, \notag\\
h_2(r_i, t_i,\beta_i;\theta) &= \left|\left|\min\{D_tV_\theta(r_i,t_i;\beta_i) + H(r_i,\nabla V_\theta,t_i;\beta_i),\, l(r_i) - V_\theta(r_i,t_i;\beta_i)\}\right|\right|, \notag\\
h(r_i,t_i,\beta_i;\theta) &= h_1(r_i,t_i,\beta_i;\theta)+\lambda h_2(r_i,t_i,\beta_i;\theta).
\end{align}
Here, each loss term represents the parameterized value function and the RHS of HJI-VI \eqref{eqn:HJIVI}. $\theta$ represents the parameters of the DNN. The term $h_1(\cdot)$ is the difference between the parameterized value function, $V_\theta(\bar \rstate_i,t_i;\beta_i)$, and the boundary condition, $\ell(\rstate_i)$, which ultimately represents the ground truth value function. This allows the DNN to propogate the value function backwards during training. $h_2(\cdot)$ guides the DNN to train the parameterized value function consistent with the HJI-VI, which is necessary since true solutions are derived from it. Finally, $h(\cdot)$ weighs $h_1(\cdot)$ and $h_2(\cdot)$ via $\lambda$ to determine the importance between the  ground truth value function and the HJI-VI.


 \begin{figure}
    \centering
    \includegraphics[width=.7\textwidth]{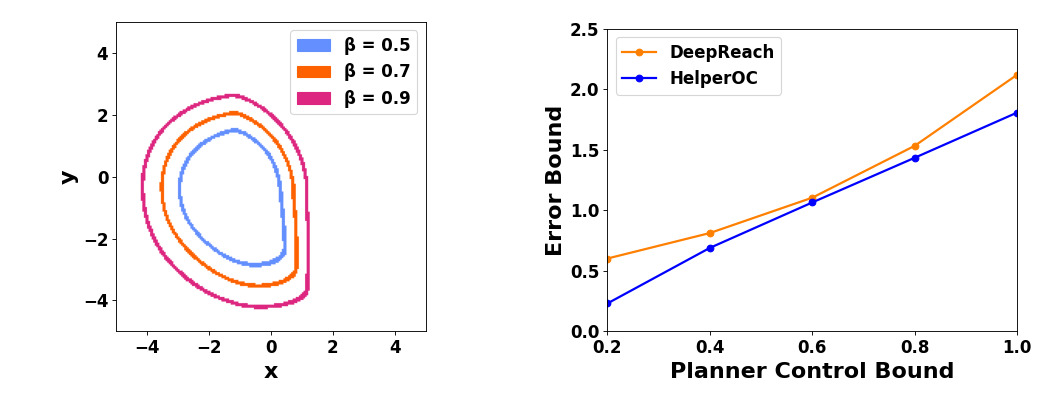}
    \caption{(Left) TEB for system \eqref{eqn: para Dubins} with different planner control bounds $\beta$. Training parameters: 40k pre-train iterations, followed by 110k training iterations. The model is trained until convergence. Total training took 10h32m on a NVIDIA A30. As $\beta$ increases, the TEB grows larger. (Right) comparison between standard dynamic programming based HJ reachability (using codebase HelperOC) and DeepReach. Empirically, DeepReach produces more conservative error bounds. \vspace{-1em}} 
    \label{fig:2}
\end{figure}

 A three-layer DNN with a hidden layer size of 512 neurons is trained using the loss function \eqref{eq:loss_function}. From $V_\theta(r,t;\beta)$, the TEB can be queried at the convergence time in the relative space. We denote the converged parameterized value function as $V_{\theta}^\infty(r;\beta)$. For simplicity, the parameter state queried will represent both $\beta_1$ and $\beta_2$, meaning $\beta_1 = \beta$ and $\beta_2 = \beta$. 
 
 Figure \ref{fig:2} (left) shows the dTEB corresponds to different $\beta$ for \eqref{eq:ex_dubins}. It can be seen that larger $\beta$ result in larger dTEB. This is because larger $\beta$ means more authority for the planner, making it harder for the tracker to track. We empirically show that the error bounds (largest euclidean distance to the boundary of the set) computed by DeepReach is conservative, also shown by Fig.~\ref{fig:2} (right). More details on determining the dTEB is explained in section 4 with \textit{queryPlannerControl} function.



\noindent \textbf{Static and Dynamic TEB.} In Parametric FaSTrack, we define two new terms for the TEB: dTEB $\TEB_{e,d}^\infty (\beta)$ and sTEB $\TEB_{e,s}^\infty$. The dTEB in the planner space is defined as the minimal level set of the value function at a particular $\beta$, i.e. 
$\TEB_{e,d}^\infty(\beta) := \{e: \exists \eta \text{ s.t. }\valfunc_\theta^\infty(e,\eta; \beta) \leq \minval_\theta ^\infty \}$. The sTEB is the error bound when the planner is navigating through the environment \textit{as slowly as possible}, i.e. with its minimum control bound :$\TEB_{e,s}^\infty := \{ e: \exists \eta \text{ s.t. }\valfunc_\theta^\infty(e,\eta; \beta_l) \leq \minval_\theta ^\infty \}$. As the robot moves in the unknown environment, its distance to the obstacle changes, and therefore different planner control bounds $\beta$ are applied. The dTEB changes corresponds to the $\beta$ applied, while sTEB stays unchanged. 
 




\noindent \textbf{Tracking Controller.} The tracking controller given the relative states can be computed as in \eqref{eq:tracking_control} from the parameterized value function evaluated at the current value of $\beta$.
\vspace{-1em}

%% file: sections/planning.tex
The planner control bounds $\beta$ can have a major influence on the speed of the navigation process for FaSTrack. 
In our approach, we adapt $\beta$ based on the distance from the autonomous system (tracker) to the obstacle while guaranteeing safety. The overall framework is described in Alg.~\ref{algo: 3} and Fig.~\ref{fig:1} (right). We additionally introduce the \textit{queryPlannerControl} function, which provides the largest $\beta$ possible for maximal efficiency, and the \textit{adjustPlannerControl} function, which intelligently reset the planning state when necessary to guarantee safety. We do not make restrictions on the planning algorithm used. To provide safety guarantees, the sensing range is assumed to be twice the value of $\dTEB$ when $\beta$ is the upper bound parameter value, denoted as $\beta_u$.

\algnewcommand{\WHILE}{\textbf{while}}
\algnewcommand{\ENDWHILE}{\textbf{end while}}
\algnewcommand{\DO}{\textbf{do}}
\algnewcommand{\IF}{\textbf{if}}
\algnewcommand{\ELIF}{\textbf{else if}}
\algnewcommand{\ENDIF}{\textbf{end if}}
\algnewcommand{\THEN}{\textbf{then}}
\algnewcommand{\ELSE}{\textbf{else}}
\begin{algorithm}[t]
\begin{multicols}{2}

\begin{algorithmic}[1]
\Require $V_{\theta}^\infty(r;\beta)$ and gradient $\nabla V_{\theta}^\infty(r;\beta)$,  $\TEB_{e,s}^\infty$, $\dTEB$, initial states $s_0$, $p_0$, $\beta_l$, $\beta_u$\\
\textbf{Initialization:}
\State Set initial state, $\dTEBmax$, $\beta_{\text{query}}$, and time: $s \leftarrow s_0$, $\dTEBmax \leftarrow \infty$, $\beta_{\text{query}} \gets \beta_u$, replanFlag=1. 
\State \WHILE \* not near goal \DO
\State \hspace{0.5cm} Sense for obstacles ($\mathcal C_{\text{p, sense}}$) \label{3rd:line:5}
\State \hspace{0.5cm} Find min distance from obstacles $\dist_{\text{obs}}(s)$
\State \hspace{0.5cm} $\beta_{\text{old}} \gets \beta_{\text{query}}$
\State \hspace{0.5cm} $\beta_{\text{query}},\dTEBmax \gets$\textit{queryPlannerControl}\label{3rd:line:8}
\State \hspace{0.5cm} Augment obstacles by $\mathcal C_{\text{p, aug}}$
\State \hspace{0.5cm} \IF \* obstacle sensed OR replanFlag=1 \THEN \label{3rd:line:10}
\State \hspace{1cm} $p_{\text{raw}} \gets$ run planner, replanFlag=0
\State \hspace{0.5cm} \ENDIF \label{3rd:line:12}
\State \hspace{0.5cm} Find closest planner state $p^*$ from $s$
\State \hspace{0.5cm} $p_{\text{next}} \gets$ \textit{adjustPlannerControl}
\State \hspace{0.5cm} Find next relative states $r_{\text{next}}$ from $p_{\text{next}}$\label{3rd:line:15}
\State \hspace{0.5cm} Obtain $s_{\text{next}}$ by applying $u_s^*$ to the tracker
\State \hspace{0.5cm} $s \leftarrow s_{\text{next}}$, $p \leftarrow p_{\text{next}}$\\
\ENDWHILE 
\State \textbf{queryPlannerControl}
\State $\beta_{\text{query}} \gets \beta_l$ \label{lst:line:3}
\State \IF \* $\TEB^\infty_{e,s} \nsubseteq ball(0,\dist_{\text{obs}}/2)$ \THEN 
\State \hspace{0.5cm} $\beta_{\text{query}} \gets \beta_l$
\State \ELSE
\State \hspace{5mm}  \WHILE \* $\TEB^\infty_{e,d}(\beta) \subset ball(0,\dist_{\text{obs}}/2)$ \DO
\State \hspace{10mm} $\beta_{\text{query}y} \gets \beta_{\text{query}} + \Delta\beta$
\State \hspace{5mm}\ENDWHILE
\State \ENDIF
\State $\beta_{\text{query}} \gets \beta_{\text{query}} - \Delta\beta$ \label{lst:line:10}
\State $\dTEBmax \gets \TEB^\infty_{e,d}(\beta_{\text{query}})$
\State Return $\beta_{\text{query}}$, $\dTEBmax$ \label{lst:line:12}

\State \textbf{adjustPlannerControl}
\State \IF \* $\beta_{\text{query}} < \beta_{\text{old}}$ \THEN 
\State \hspace{0.5cm} \IF \* $s-Qp^* \in \dTEBmax$ \THEN \label{2nd:line:2}
\State \hspace{1cm} $p_{\text{next}} \gets p^*$ \label{2nd:line:3}
\State\hspace{1cm} replanFlag $\gets 0$ \label{2nd:line:4}
\State \hspace{0.5cm} \ELIF \* $s-Qp^* \notin \dTEBmax$ \THEN \label{2nd:line:5}
\State \hspace{1cm} $p_{\text{next}} \gets$ a state $p$ in the $\mathcal C_{\text{p,aug}}$ free region such that $s-Qp \in \dTEBmax$ 
\State \hspace{1cm} 
replanFlag $\gets 1$ \label{2nd:line:7}
\State \hspace{0.5cm} \ENDIF
\State  \ELIF \* $u_{p,\text{old}} \leq \beta_{\text{query}}$ \THEN \label{2nd:line:9}
\State \hspace{0.5cm}$p_{\text{next}}\gets p\in p_{\text{raw}}$ discretized by $\beta_{\text{query}}$ 
\State \ENDIF \label{2nd:line:11}
\State $p_{\text{next}}$ is removed from $p_{\text{raw}}$
\State Return $p_{\text{next}}$

\end{algorithmic} 
\end{multicols}
\caption{Online Parametric FaSTrack (converged value function)}\label{algo: 3}
\end{algorithm}

At runtime, the autonomous system first senses the environment (line \ref{3rd:line:5} of Alg.~\ref{algo: 3}, top of Fig.~\ref{fig:1}). The obstacles (if sensed) are augmented with sTEB in the planning space, denoted as $\mathcal C_{\text{p, aug}}$. Any path planning algorithm may now be used to generate a raw path, which is a set of planner states, denoted as $p_{\text{raw}} $. This path is guaranteed to be free of augmented obstacles. Note that a new raw path is generated if a new obstacle is sensed or the replanFlag is 1 (line \ref{3rd:line:10}-\ref{3rd:line:12} of Alg.~\ref{algo: 3}.) 

Next, the planner moves along the raw path based on its planner control bound $\beta$. The appropriate $\beta$ must be determined such that the planner moves as quickly as possible while maintaining safety.
First, the minimum distance between the tracker and the every sensed obstacle is computed, which is denoted as $\dist_{\text{obs}}(s)$, and $\dist_{\text{obs}}(s) = \min_{ b \in \mathcal C} || s - b||$.
The largest acceptable planner control bound $\beta_{\text{query}}$ is determined by $\dist_{\text{obs}}(s)$, and the corresponding dTEB is denoted as $\dTEBmax$ (left branch of Fig.~\ref{fig:1}). 
When $\TEB^\infty_{e,s} \nsubseteq ball(0,\dist_{\text{obs}}/2)$, $\dTEBmax = \TEB^\infty_{e,s}$, otherwise $\dTEBmax$ is the largest dTEB contained in the ball with radius $\dist_{\text{obs}}/2$ and centered at the origin, $ball(0,\dist_{\text{obs}}/2)$ (line \ref{lst:line:3}-\ref{lst:line:12} of Alg.~\ref{algo: 3}). 

\begin{remark} 
    The \textit{queryPlannerVelocity} function constrains the norm of the relative state after planning and before the tracking. If $\TEB^\infty_{e,d}(\beta) \subset ball(0,\dist_{\text{obs}}/2)$, we guarantee the norm is at most $\dist_{\text{obs}}/2$, and the distance between planner and obstacles is greater than $\dist_{\text{obs}}/2$. After tracking, the norm of new relative state will also be smaller than $\dist_{\text{obs}}/2$, due to the control invariance of dTEB. If $\TEB^\infty_{e,d}(\beta) \nsubseteq ball(0,\dist_{\text{obs}}/2)$, we guarantee that the relative state stays in the sTEB, while the distance between planner and obstacles is greater than the radius of the sTEB.  
\end{remark}
\vspace{-0.5em}

\begin{figure}[t]
    \centering\includegraphics[width=1\textwidth]{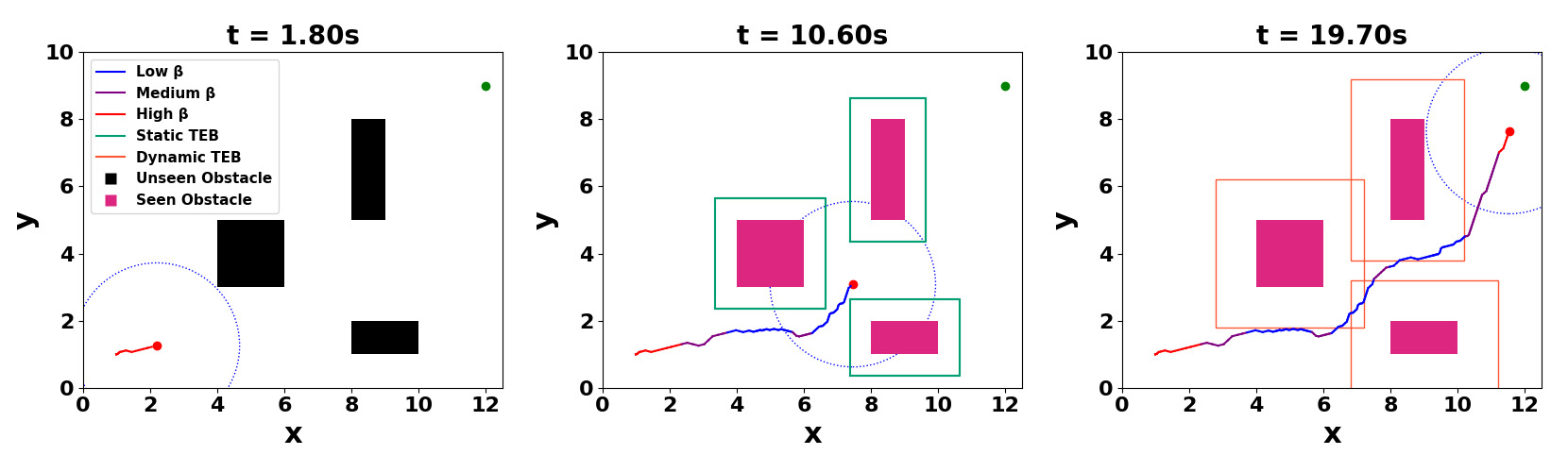}
    \caption{Online simulation of 6D Dubin's car. The trajectory is color-coded with the speed of planning: low $\beta = 0.5$ (blue), medium $0.5 < \beta \leq 1$ (purple), and high $\beta > 1$ (red). The goal (green dot) and current tracker state (red dot) are shown. The sTEB is in green. The dTEB for the current $\beta$ is in orange. The dotted circle around the tracker represents the sensing range. The left panel shows the planner applying $\beta_u$ at the initial time since no obstacles are sensed. In the second panel, the system senses an obstacle ($\dTEBmax \subseteq \TEB_{e,s}^\infty$), and slows down to reduce the corresponding TEB. In the third panel, the system applies higher $\beta$ values as it moves away from the obstacles and towards the goal.\vspace{-.5em}}
    \label{fig:3}
\end{figure}

\begin{wrapfigure}{r}{0.35\textwidth}
  \begin{center}
    \includegraphics[width=0.35\textwidth]{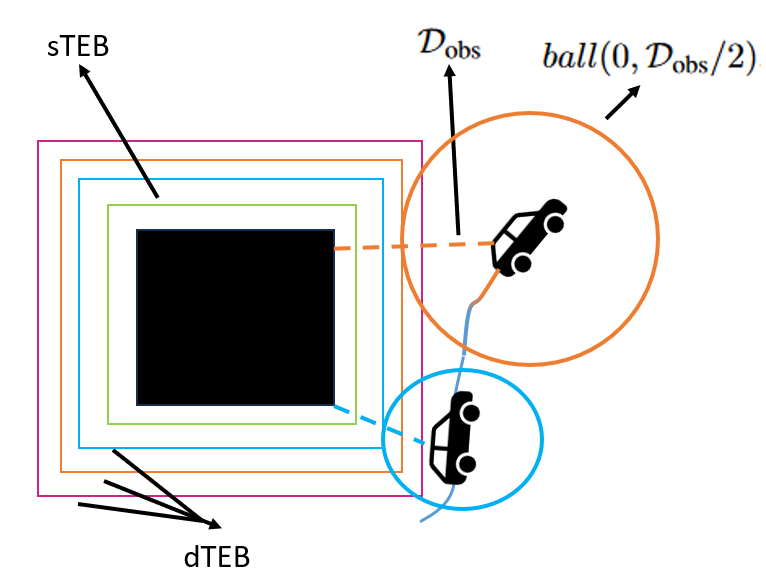}
  \end{center}
  \caption{Tracker moving away from the obstacle. $\dTEBmax$ expands and safety is preserved.} 
\label{fig:moving away}

\end{wrapfigure}








The last iteration's planner control bound is denoted $\beta_{\text{old}}$, and the closest planner state from $s$ to $p_{\text{raw}}$ states is $p^*$. Given $p_{\text{raw}}$, $\beta_{\text{query}}$, $\beta_{\text{old}}$, $p^*$, and $\dTEBmax$, the \textit{adjustPlannerControl} function adapts $\beta_{\text{query}}$. There are two cases considered: switching from a larger bound to a smaller one $\beta_{\text{query}} < \beta_{\text{old}}$ and vice versa. 
If $\beta_{\text{old}} \leq \beta_{\text{query}}$, this means $\dTEBmax$ has increased and the tracker is moving away from the obstacle, therefore increasing $u_p$ won't jeopardize safety (Fig. \ref{fig:moving away}). The next plan state is determined by discretizing $p_{\text{raw}}$ with $\beta_{\text{query}}$ (Alg.~\ref{algo: 3}, lines \ref{2nd:line:9}-\ref{2nd:line:11}).

On the contrary, if $\beta_{\text{query}}<\beta_{\text{old}}$, this means the tracker is moving towards an obstacle and therefore $\dTEBmax$ is shrinking compared to last iteration. In this case, we must ensure that the true system (i.e. the tracker) maintains safety in the smaller TEB. In other words, the relative state ($s - Qp^*$) must be contained in $\dTEBmax$. If so, $p^*$ is a planner state that satisfies the safety constraints (line \ref{2nd:line:2}-\ref{2nd:line:4} of Alg.~\ref{algo: 3}). 
If the relative state is not contained in $\dTEBmax$, there is no planner state that satisfies the safety constraints on the raw path, and the planner state must be reset. However, we cannot simply reset the planner state to the tracker state, as the tracker state may be in the sTEB-augmented obstacle region. The reset planner state must be in the sTEB-augmented obstacle free region (line \ref{2nd:line:5}-\ref{2nd:line:7} of Alg.~ \ref{algo: 3}). To do this, we select a random point that is in $\dTEBmax$ centered at current tracker state and not in $\mathcal C_{\text{p,aug}}$.

After adjusting the planner control bound, $p_{\text{next}}$ is removed from $p_{\text{raw}}$, and is sent to the tracking system for finding the optimal safe controller. This process is identical to FaSTrack: compute the relative state after planning system moves, find the optimal control $u_s$ from $\nabla V_{\theta}^\infty(r;\beta_{\text{query}})$, and apply $u_s$ for $\Delta t$. The planner updates its next state, and the algorithm iterates.

%% file: sections/results.tex
We validate our method with a 6D Dubin's car system and a 13D quadcopter system.

\begin{figure}[t]
\begin{minipage}{0.375\textwidth}\centering
    \includegraphics[width=1\textwidth]{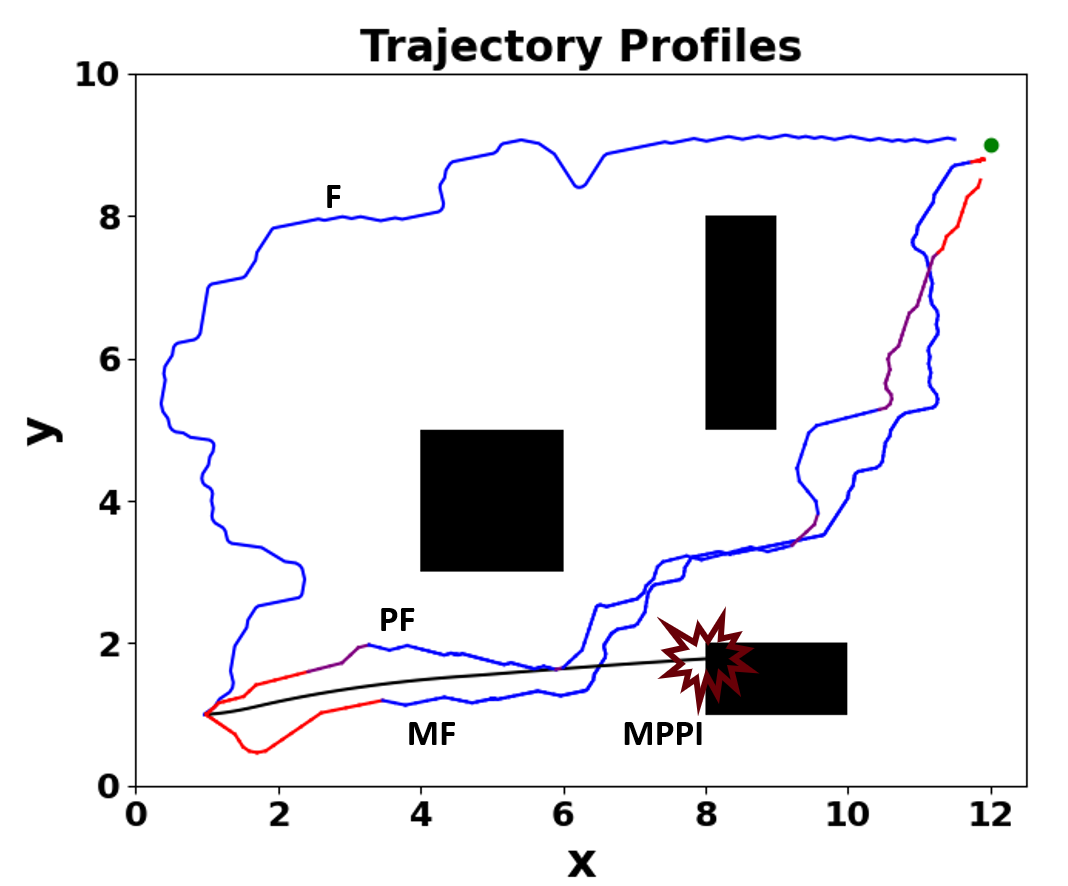}
    \caption{Trajectory for F, MF PF, and MPPI. Trajectory colors correspond to Fig.~\ref{fig:3}. MPPI does not have a color since it does not have a $\beta$.}
    \label{fig:4}
\end{minipage}\hfill
\begin{minipage}{0.6\textwidth}
    \small
    \vspace{1cm}
    \begin{tabular}{ |p{3.15cm}||p{0.7cm}|p{0.7cm}|p{1.4cm}|p{0.8cm}| }
    \hline
    \multicolumn{5}{|c|}{Trajectory Planner Benchmark (20 Run Average)} \\
    \hline
    Metrics& F & MF & \textbf{PF (ours)} & MPPI \\
    \hline
    Reached Goal (\%) & 100 & 100 &  100 & 65\\
    Obstacle Collision (\%)&  0  & 0   & 0 & 35\\
    Solution Time (s)& 37.31 & 30.11 &  22.47 & 10.94\\
    \hline
    \end{tabular}
\captionof{table}{Benchmark testing of trajectory planners; each row is averaged across 20 runs. F: FaSTrack; MF: Meta-FaSTrack; PF: Parametric FaSTrack; MPPI: Model Predictive Path Integral.\label{tab:1}}
    \end{minipage}
    \vspace{-1em}
\end{figure}
\subsection{6D Dubin's Car System}
Given the running example model \eqref{eqn: para Dubins}, the online execution of parametric FaSTrack is shown  in Fig.~\ref{fig:3}. It can be seen that the planner control bound changes as the distance to the obstacle changes. 
We performed a benchmark test in Table \ref{tab:1} on four different trajectory planners to compare safety, speed, and feasibility metrics. Feasibility is satisfied if the tracker reaches the goal, safety is achieved if there are no collisions, and solution time is the time it takes for the tracker to reach the goal. All of the testing was done on identical obstacle maps, control inputs, and $\beta$ (if applicable):\\
\noindent \textbf{Feasibility.} All of the reachability based trajectory planner reaches its goal 100\% of the time, while MPPI only reaches its goal 65\% of the time due to collisions.\\
\textbf{Safety.} All of the reachability based trajectory planners preserve safety over 20 runs, while MPPI collides with obstacles 35\% of the time. An example of collision for MPPI is shown in Fig.~\ref{fig:4}, since there are no safety guarantees for MPPI. The reachability planners are able to stay relatively distant from the obstacles, albeit trading off navigation speed. \\
\textbf{Solution Time.} MPPI was faster than all of the reachability based trajectory planners on average because MPPI tracking does not account for safety, thus speed can be prioritized. For the reachability based methods, parametric FaSTrack was the fastest, averaging 22.47s. Fig.~\ref{fig:4} shows the trajectory profiles of each framework. FaSTrack performed the slowest since it had to plan around the obstacles to reach the goal. Although Meta-FaSTrack had a similar trajectory profile to Parametric FaSTrack, PF is able to adapt to its environment every iteration, which resulted in a faster solution time.

\subsection{13D Quadcopter System}

\begin{figure}[t]
    \centering
    \includegraphics[width=1\textwidth]{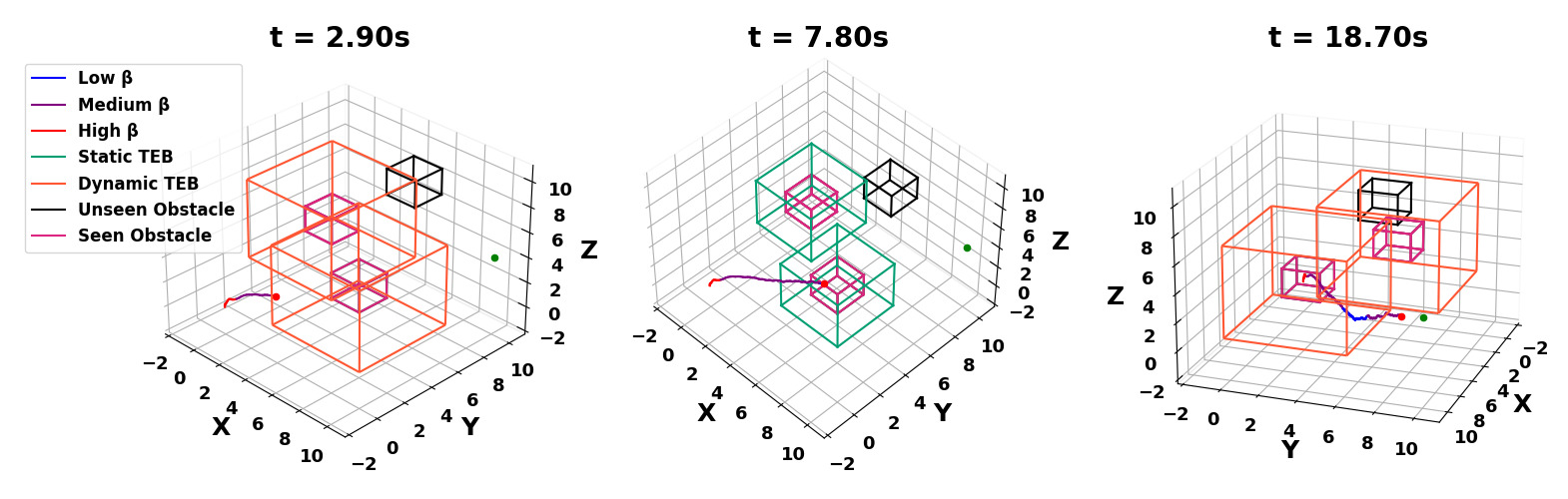}
    \caption{Online simulation of 13D quadcopter relative system. The color set up is the same as in Figure \ref{fig:3}
    Faster planning speed leads to larger error bounds, and thus can only be applied in open environments. The framework applies higher planner control bounds in relatively open environments and lower bounds in tight environments. The system did not raise the replanFlag during online planning, which implies that the system is able to track the planner despite adjusting $\beta$.\vspace{-.5em}}
    \label{fig:5}
\end{figure}

For simplicity, we directly provide the parameterized relative system with 3 virtual states added:
\begin{align*}
    \dot  \rstate_1 = \rstate_2 - u_{px}, \hspace{2mm}
    \dot  \rstate_2 = g\tan(\rstate_3), \hspace{2mm}
    \dot  \rstate_3 = -d_1\rstate_3 + \rstate_4, \hspace{2mm}
    \dot  \rstate_4 = -d_0\rstate_3 + n_0 u_x, \hspace{2mm}\\
    \dot  \rstate_5 = \rstate_6 - u_{py}, \hspace{2mm}
    \dot  \rstate_6 = g\tan(\rstate_7), \hspace{2mm}
    \dot  \rstate_7 = -d_1\rstate_7 + \rstate_8, \hspace{2mm}
    \dot  \rstate_8 = -d_0\rstate_7 + n_0 u_y, \hspace{2mm}\\
    \dot  \rstate_9 = s_{10} - u_{pz}, \hspace{2mm}
    \dot  \rstate_{10} = k_T u_z - g, \hspace{2mm}
    \dot  \beta_{1} = 0, \hspace{2mm}
    \dot  \beta_{2} = 0, \hspace{2mm}
    \dot  \beta_{3} = 0, 
\end{align*}
where $u_x,u_y,u_z$ are tracker controls, $u_{px},u_{py},u_{pz}$ are planner controls. $g = 9.81$, $n_0 = 10$,  $d_1 = 8$, $d_0 = 10$, and $k_T = 0.91$, $\left|u_x\right|, \left|u_y\right| \leq \pi/9$, $u_z \in [0, 1.5g]$, and $\beta_1,\beta_2,\beta_3 \in [0.5 , 1.5]$.
The parameterized relative system was trained for a total of 14h10m on a NVIDIA A30, with 50k pretrain iterations and 150k training iterations. Figure \ref{fig:5} depicts the online simulation results in 3D space. Similar to the Dubin's car example, the quadcopter initially applies $\beta_u$ at runtime, but slows down near obstacles and applies $\beta_l$. Near the end of the time horizon, the quadcopter applies higher $\beta$ as it moves away from obstacles. 

%% file: sections/conclusions.tex
In this work, we introduced a deep learning based modification to FaSTrack, which leverages online adaptability and high dimensional feasibility of DeepReach for fast and safe online planning. Our framework naturally trades off the planning speed with safety to move quickly through open environments and carefully through cluttered environments. Empirical simulation results show that our method greatly increased the navigation speed up to 40\% compared to original FaSTrack work while preserving safety. 
Future directions include providing formal probabilistic guarantees for the trained value function (\cite{10160600}), proving guaranteed conservative error bounds for certain planning model formulations (\cite{8793919}), adaptation to dynamic obstacles (\cite{fisac2018probabilistically}), smoothing the tracker trajectory, and an extension to other sensors like camera. 




